\documentclass[10pt,twocolumn,letterpaper]{article}

\usepackage[pagenumbers]{cvpr} 

\usepackage{graphicx}
\usepackage{amsmath}
\usepackage{amssymb}
\usepackage{booktabs}
\usepackage{bm}
\usepackage{xcolor}
\usepackage[normalem]{ulem}
\usepackage{enumitem}
\usepackage[accsupp]{axessibility}

\usepackage[pagebackref,breaklinks,colorlinks]{hyperref}

\usepackage[capitalize]{cleveref}
\crefname{section}{Sec.}{Secs.}
\Crefname{section}{Section}{Sections}
\Crefname{table}{Table}{Tables}
\crefname{table}{Tab.}{Tabs.}

\def\etal{\textit{et~al.~}}		  

\def\sysname{NeRFReN}

\def\naive{na{\"i}ve\xspace}

\DeclareMathAlphabet{\altmathcal}{OMS}{cmsy}{m}{n}

\newlength\paramargin
\newlength\figmargin

\newlength\secmargin
\newlength\figcapmargin
\newlength\tabcapmargin

\newcommand{\topic}[1]
{
\vspace{1.5mm}\noindent\textbf{#1}
}

\long\def\ignorethis#1{}

\newbox\jsavebox%

\makeatletter
\newcommand{\providelength}[1]{%
  \@ifundefined{\expandafter\@gobble\string#1}
   {
    \typeout{\string\providelength: making new length \string#1}%
    \newlength{#1}%
   }
   {
    \sdaau@checkforlength{#1}%
   }%
}

\def\cvprPaperID{3703} 
\def\confName{CVPR}
\def\confYear{2022}

\begin{document}

\title{\sysname: Neural Radiance Fields with Reflections}
\author{
\vspace{2mm}
Yuan-Chen Guo$^{1}$\thanks{Research done when Yuan-Chen Guo was an intern at Tencent AI Lab.} \quad
Di Kang$^2$ \quad
Linchao Bao$^2$ \quad
Yu He$^3$\thanks{Corresponding author.} \quad
Song-Hai Zhang$^1$
\\ $^1$BNRist, Department of Computer Science and Technology, Tsinghua University, Beijing
\\ $^2$Tencent AI Lab \quad $^3$BIMSA
\\ {\tt\small guoyc19@mails.tsinghua.edu.cn \quad \{dkang,linchaobao\}@tencent.com}
\\ {\tt\small hooyeeevan2511@gmail.com \quad shz@tsinghua.edu.cn}
}
\maketitle

\begin{abstract}
Neural Radiance Fields (NeRF) has achieved unprecedented view synthesis quality using coordinate-based neural scene representations. 
However, NeRF's view dependency can only handle simple reflections like highlights but cannot deal with complex reflections such as those from glass and mirrors. 
In these scenarios, NeRF models the virtual image as real geometries which leads to inaccurate depth estimation, and produces blurry renderings when the multi-view consistency is violated as the reflected objects may only be seen under some of the viewpoints. 
To overcome these issues, we introduce NeRFReN, which is built upon NeRF to model scenes with reflections. 
Specifically, we propose to split a scene into transmitted and reflected components, and model the two components with separate neural radiance fields. 
Considering that this decomposition is highly under-constrained, we exploit geometric priors and apply carefully-designed training strategies to achieve reasonable decomposition results. 
Experiments on various self-captured scenes show that our method achieves high-quality novel view synthesis and physically sound depth estimation results while enabling scene editing applications. 
Project webpage: \url{https://bennyguo.github.io/nerfren/}.
\vspace{-0.5cm}
\end{abstract}

\section{Introduction}

\begin{figure}[t]
  \centering
   \includegraphics[width=1.0\linewidth]{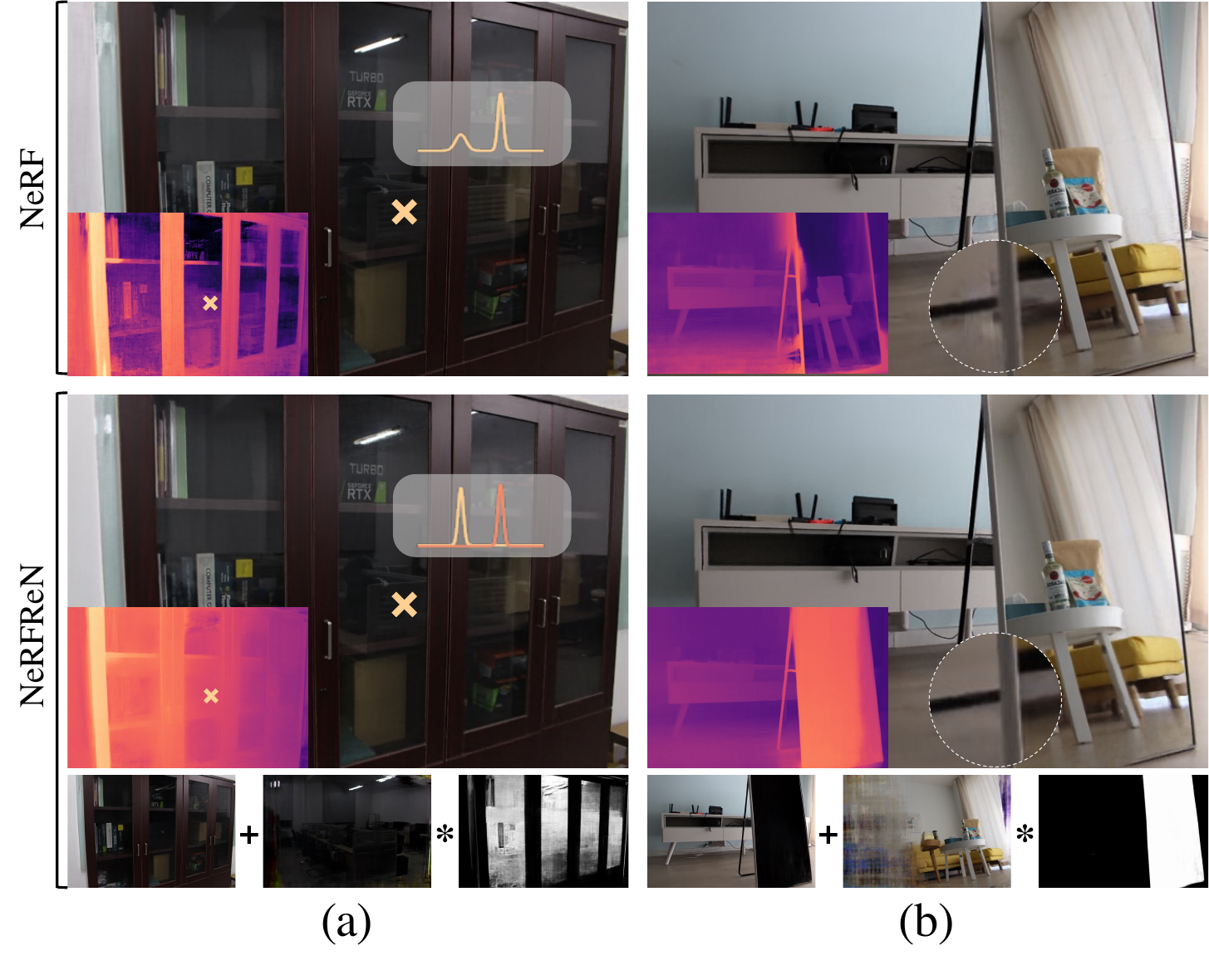}
   \caption{
   NeRF models the stable reflection image as real geometries rather than view-dependent effect of the points on the reflective surface. 
   We illustrate two problems NeRF may encounter when modeling scenes with complex reflections: (1) inaccurate depth estimation in reflective regions, as in (a) and (b); 
   (2) inaccurate rendering when the multi-view consistency is violated, as in the magnified area in (b). {\sysname} tackles these problems by modeling the transmitted and reflected component of the scene with separate NeRFs, and synthesizing views in the image domain (bottom). See \cref{sec:nerf} for more detailed analysis.}
   \label{fig:teaser}
  \vspace{-0.5cm}
\end{figure}

Photorealistic novel view synthesis (NVS) from unstructured image collections is crucial to creating immersive virtual experiences.
Despite having achieved significant progress in controlled settings, challenges still exist in handling light transport at the surface of different materials. 
For example, reflections caused by glass or mirrors commonly exist in real-world scenes, posing great difficulties for novel view synthesis due to the severe view-dependent effects.

Neural Radiance Fields (NeRF)~\cite{mildenhall2020nerf}, as an emerging technique in this field, has achieved impressive view synthesis quality by adopting volumetric representations with coordinate-based neural networks. 
By conditioning radiance on both coordinates and viewing directions, NeRF is able to faithfully handle view-dependent effects like highlights. 
However, in the presence of severe reflections that contain a stable virtual image other than highlights, NeRF tends to model the scene geometry behind the reflective surface as translucent (like fogs) and regard the reflected geometries as appearing at some virtual depth. 
The reason is that NeRF does not explain the view-dependency of a reflecting surface point by changing the color of this very point when viewed from different directions, but explain it by utilizing all the spatial points behind it along the camera ray to get the correct color through volume rendering as in~\cref{eqn:volume-rendering}, resulting in foggy geometries and physically wrong depth.
Although this works well for view synthesis in certain scenarios, there are 
two inherent limitations: 
(1) it is hard to estimate the correct scene geometry, preventing it from further explanations or editing; 
(2) inaccurate renderings are generated when the multi-view consistency does not hold since the reflected objects may only be seen in some of the images, which limits the views NeRF can correctly synthesize.

To this end, we propose {\sysname}, which enhances \textbf{NeRF} to handle scenes with \underline{\textbf{Re}}flectio\underline{\textbf{N}}s.
Instead of representing the whole scene with a single neural radiance field, we propose to model the \emph{transmitted} and \emph{reflected} parts of the scene with separate neural radiance fields. 
To synthesize novel views, the transmitted image $I_t$ and reflected image $I_r$ rendered by the corresponding radiance fields are composed in an additive way, where the reflected image $I_r$ is weighted by a learned \emph{reflection fraction map} $\bm{\beta}$:
\begin{equation}\label{eqn:formulation-image}
I = I_t + \bm{\beta} I_r
\end{equation}

However, it is highly under-constrained to decompose the scene into transmitted and reflected components in an unsupervised way.
There are infinitely many solutions that could lead to correct renderings on training images. 
Common degradations are 
(1) having one of the components explain the full scene while leaving the other empty; 
(2) two components both model the full scene;
(3) somewhere between (1) and (2), where part of the one component is mixed into the other one, resulting in incomplete transmitted/reflected geometry. 
To solve the ambiguities, we made the following assumptions:
\begin{itemize}[align=parleft,left=0pt..1em]
    \itemsep 0em 
    \item \textbf{Assumption1:} the reflection fraction is only related to the transmitted component, since it indicates the material of the reflecting surface;
    \item \textbf{Assumption2:} the transmitted component has a locally smooth depth map, since most of the reflectors in real world scenes are almost planar;
    \item \textbf{Assumption3:} the reflected component requires only simple geometry to present correct renderings since most of the time we can only see the reflection image from very limited viewing directions.
\end{itemize}
These assumptions generally hold true for planar or almost planar reflective surfaces commonly seen in real-world scenes. Accordingly, we design a specific network architecture for Assumption1, apply a depth smoothness prior for Assumption2, and a novel bidirectional depth consistency constraint for Assumption3.
Together with our warm-up training strategy, we are able to faithfully decompose the transmitted and reflected components, achieving promising novel synthesis results on various real-world scenes.
As for even more challenging cases (such as mirrors) in which the ambiguities cannot be solved by the specific design, we exploit an interactive setting where a small number of user-provided reflection masks can be utilized to get the correct decomposition (\cref{fig:decomposition}, tv and mirror).

To summarize, our contributions are:
\begin{itemize}[align=parleft,left=0pt..1em]
    \itemsep 0em 
    \item We analyze NeRF's behavior and limitations in modeling reflective regions, and propose to use separate transmitted and reflected neural radiance fields for scenes with complex reflections.
    \item We devise specific network architectures, exploit geometric priors, and apply carefully-designed warm-up training strategy, to achieve physically sound decomposition results with competitive novel view synthesis quality compared to vanilla NeRF on several real world scenes, including challenging cases like scenes with mirrors.
    \item We also investigate depth estimation and scene editing as further applications based on the decomposed results.
\end{itemize}

\section{Related Work}

\topic{Neural scene representations for novel view synthesis.}
Novel view synthesis aims to generate images observed from novel views based on a set of captured images of the scene. 
In order to get correct synthesis results, the underlying 3D geometry of the scene must be taken into consideration, for which various scene representations are proposed. 
Image-based rendering~\cite{buehler2001unstructured,snavely2006photo,hedman2016scalable,riegler2020free,riegler2021stable,thies2019deferred} mainly utilizes a mesh model of the scene, which is usually reconstructed from offline Structure-from-Motion (SfM) and Multi-View-Stereo (MVS) methods~\cite{schoenberger2016sfm,schoenberger2016mvs}. 
Other commonly adopted representations include voxel grid~\cite{sitzmann2019deepvoxels}, volume~\cite{lombardi2019neural}, point cloud~\cite{aliev2020neural} and multi-plane images~\cite{flynn2019deepview,wizadwongsa2021nex}.

In recent years, coordinate-based neural representations~\cite{mescheder2019occupancy,chen2019learning,park2019deepsdf,sitzmann2019scene,mildenhall2020nerf} have shown remarkable capability of modeling the 3D world. 
Among them, Neural Radiance Fields (NeRF)~\cite{mildenhall2020nerf}, as an emerging technique in the novel view synthesis domain, models the scene as a continuous volumetric field parameterized by neural networks. 
NeRF produces astonishing novel view synthesis results in several casually-captured real scenes, even when view-dependent effects like highlights present. 
Our work extends NeRF to support view synthesis in scenes with reflective surfaces like glass and mirrors, which are common in real world scenes and strongly affect the immersion of virtual experiences.

\topic{Reflections in rendering.}
Reflections are common in real world, but can be hard to correctly model without simulating the light paths using techniques like ray-tracing. 
In computer graphics, Screen Space Reflection (SSR) technique is developed to simulate the reflection effects with low costs. 
Synthesizing novel views in scenes with reflections has long been a challenging problem, but is rarely explored~\cite{sinha2012image,xu2021scalable}. 
Sinha \etal~\cite{sinha2012image} for the first time targets at novel view synthesis in scenes with reflections using image-based rendering. 
They propose to decompose each image into a transmitted layer and a reflected layer combined with a binary reflection mask. 
They approximate the geometry of each layer as piece-wise planar surfaces, and estimate their depth by two-layer stereo matching. 
The reflection mask is optimized using graph cut based on depth uncertainty. 

Our work adopts similar image formulation as Sinha \etal~\cite{sinha2012image}, but models the scene with NeRF for its state-of-the-art performance on the novel view synthesis task. 
NeRF resolves view-dependency by taking the viewing direction as network input, but is only able to model low-frequency view-dependent effects. 
We extend NeRF by modeling the transmitted and reflected components with two separate neural radiance fields, and achieve visually appealing results in scenes with complex reflections.

Our work is also different from existing NeRF-based inverse rendering methods~\cite{boss2021nerd,zhang2021nerfactor}, which assuming opaque surfaces with simple BRDF, distant environment lighting and local illumination models. These assumptions are not suitable in our case with complex reflections. In contrast, we assume composition of transmitted and reflected components in an additive way.

\section{Method}

Here we present NeRFReN, a neural radiance field approach for novel view synthesis of scenes with reflections. 
We first briefly recap Neural Radiance Fields (NeRF)~\cite{mildenhall2020nerf} in \cref{sec:nerf}, and present our scene formulation and network architecture in \cref{sec:formulation}. 
To faithfully decompose the scene into transmitted and reflected components, we exploit geometric priors (\cref{sec:geometric}) and apply specifically designed training strategies (\cref{sec:training}). 
Training with minimum user input to handle hard cases is described in \cref{sec:interactive}.

\subsection{Neural Radiance Fields Revisited}\label{sec:nerf}

Neural Radiance Fields (NeRF) represents a scene as a continuous volumetric field, where the density $\sigma \in \mathbb{R}$ %
and radiance $\mathbf{c} \in \mathbb{R}^3$ %
at any 3D position $\mathbf{x} \in \mathbb{R}^3$ %
under viewing direction $\mathbf{d} \in \mathbb{R}^2$ %
are modeled by a multi-layer perceptron (MLP) 
$f_{\theta}: (\mathbf{x}, \mathbf{d}) \rightarrow (\mathbf{c}, \sigma)$,
with $\theta$ as learnable parameters. 
To render a pixel, the MLP first evaluates points sampled from the camera ray $\mathbf{r} = \mathbf{o} + t\mathbf{d}$ to get their densities and radiance, 
and then the color $\mathbf{C}(\mathbf{r})$ is estimated by volume rendering equation approximated using quadrature~\cite{max1995optical}:
\begin{equation}\label{eqn:volume-rendering}
\widehat{\mathbf{C}}(\mathbf{r};\sigma,\mathbf{c})=\sum_{k}T_i(\sigma)(1-\text{exp}(-\sigma_i\delta_i))\mathbf{c}_i
\end{equation}
where $\delta_i=t_{i+1}-t_i$ and $T_i(\sigma)=\text{exp}(-\sum_{j<i}\sigma_j\delta_j)$. $\widehat{\mathbf{C}}$ is conditioned on $\sigma, \mathbf{c}$, and $T$ is conditioned on $\sigma$ to simplify follow-up descriptions. 
We denote the contribution of a point to the cumulative color as its weight $\omega_i$:
\begin{equation}\label{eqn:point-weight}
    \omega_i=T_i(\sigma)(1-\text{exp}(-\sigma_i\delta_i))
\end{equation}
NeRF is optimized by minimizing the photometric loss:
\begin{equation}
\mathcal{L}_{pm}=||\widehat{\mathbf{C}}-\mathbf{C}||_2
\end{equation}
The depth value $t^{*}$ along the ray can be estimated by computing the expected termination depth~\cite{zhang2021nerfactor}:
\begin{equation}\label{eqn:depth}
t^{*}(\mathbf{r};\sigma)=\sum_{k}\omega_i t_i=\sum_{k}T_i(\sigma)(1-\text{exp}(-\sigma_i\delta_i))t_i
\end{equation}

NeRF models view-dependent effects by taking viewing direction $\mathbf{d}$ as the network input. 
But as will be demonstrated in~\cref{tab:compare}, the incorporation of $\mathbf{d}$ only enables NeRF to represent low-frequency view-dependent effects.
Directly applying NeRF on scenes with severe reflections leads to a mixed geometry containing both the transmitted part and the reflected part of the scene, where the former is modeled as \emph{translucent} to get correct view reconstructions. As illustrated in the first row of~\cref{fig:teaser}, the books in the bookcase appear to be ``foggy", and the mirror surface is completely transparent.

The drawbacks are two-folds.
First, depth estimation obtained by \cref{eqn:depth} can be highly inaccurate, staying somewhere between the real depth of the reflective surface and the virtual depth of the reflected image due to the existence of translucent geometries. In the gray box of~\cref{fig:teaser}(a) we visualize the weights of points along the ray corresponding to the yellow cross. The weight distribution of NeRF appears to be bi-modal, with the left peak contributed by the real surface points and right peak by the virtual surface points. The expected termination depth along this ray is biased towards the virtual depth, visualized as a darker color in the depth map on the bottom left.
Second, it is hard to get correct view synthesis results when the multi-view consistency is violated. We illustrate this in~\cref{fig:teaser}(b), where some of the reflected objects can only be seen when front-facing the mirror. But as NeRF models the reflected objects as real geometries, we can still observe foggy contents seeing from the side of the mirror as shown in the magnified area.

\begin{figure}[t]
  \centering
   \includegraphics[width=0.8\linewidth]{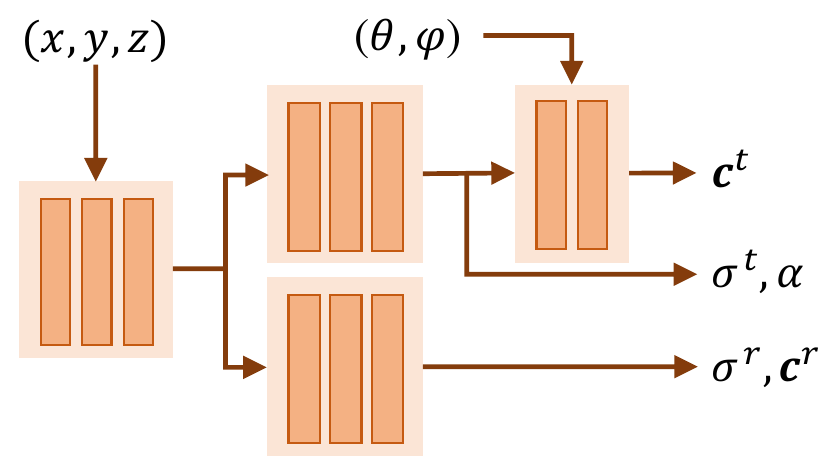}
   \caption{Network architecture of {\sysname}. The transmitted and reflected properties are predicted by separate branches of the network. Note that the reflection fraction value $\alpha$ is predicted from the transmitted branch.}
   \label{fig:network}
   \vspace{-0.5cm}
\end{figure}

\subsection{\sysname}\label{sec:formulation}

To better deal with scenes containing reflections, we propose to decompose a scene into a transmitted NeRF and a reflected NeRF.
The transmitted field has density $\sigma^t$ and radiance $\mathbf{c}^t$, and the reflected field has density $\sigma^f$ and radiance $\mathbf{c}^f$. 
A reflection fraction value $\alpha$ is learned for each 3D position to measure the reflection property of objects in different materials.
To render a pixel along ray $\mathbf{r}$, we first render the two fields respectively to get $\widehat{\mathbf{C}}(\mathbf{r};\sigma^t,\mathbf{c}^t)$ and $\widehat{\mathbf{C}}(\mathbf{r};\sigma^r,\mathbf{c}^r)$, where $\widehat{\mathbf{C}}$ is defined in \cref{eqn:volume-rendering}. 
The reflection fraction $\beta$ corresponding to the pixel is accumulated via volume rendering based on the geometry of the transmitted part:
\begin{equation}
\beta(\mathbf{r};\sigma,\alpha)=\sum_{k}T_i(\sigma^t)(1-\text{exp}(-\sigma^t_i\delta_i))\alpha_i
\end{equation}
The reflected color is attenuated by $\beta$ and composed with the transmitted color additively to get the final pixel color:
\begin{equation}\label{eqn:formulation}
\widehat{\mathbf{C}}=\widehat{\mathbf{C}}(\mathbf{r};\sigma^t,\mathbf{c}^t)+\beta(\mathbf{r};\sigma^t,\alpha)\widehat{\mathbf{C}}(\mathbf{r};\sigma^r,\mathbf{c}^r)
\end{equation}

\cref{fig:network} shows the network architecture of {\sysname}. Detailed network configurations can be found in the supplementary material.
Note that the reflection fraction value $\alpha$ is predicted along with transmitted density in the transmitted branch as it is a property of the reflective surface, uncorrelated with the reflected component.
Also, the transmitted color is conditioned on viewing direction in our design, so that the low-frequency view-dependent effects (e.g. highlights) are modeled by the transmitted field, and the stable virtual image is modeled by the reflected field.
The reflected color and the reflection fraction value are not conditioned on viewing direction. These assumptions approximate most scenes reasonably well in our experiments due to their validity\footnote{These assumptions may not hold if there exist reflectors in the reflected objects or due to severe Fresnel Effect.} and greatly reduce network complexity, making the ill-posed decomposition problem much easier to learn.

\subsection{Geometric Priors}\label{sec:geometric}
Decomposing the scene into transmitted and reflected components is an under-constrained problem. 
There are infinite number of solutions and bad local minima that may produce visually pleasing rendering results on the training images but fail to separate apart the reflected radiance field from the transmitted radiance field. 
Humans identify the reflected virtual image correctly because we are aware of the real-world geometry.
Inspired by this, we propose to utilize two geometric priors, namely a depth smoothness prior and a bidirectional depth consistency (BDC) prior, to guide the decomposition of the scene.

\paragraph{Depth smoothness.}
We exploit a general prior that the depth map of the \emph{transmitted component} should be locally smooth. Specifically, we apply the following regularization:
\begin{equation}
\begin{split}
\mathcal{L}_{d}=\sum_{p}\sum_{q\in\mathcal{N}(p)}\omega(p,q)||t^*(p)-t^*(q)||_1,
\\
\omega(p,q)=\text{exp}(-\gamma||\mathbf{C}(p)-\mathbf{C}(q)||_1)
\end{split}
\end{equation}
where $t^*(p)$ is the approximated depth defined in \cref{eqn:depth}, $p$ denotes each pixel in an image, $\mathcal{N}(p)$ is the set of its 8-connected neighbors, $\mathbf{C}$ is the image color and $\gamma$ is a hyperparameter.
$\omega(p,q)$ is a decay factor to re-weight (relax) the constraint based on the color gradient since depth discontinuities are often accompanied with abrupt color changes. In this way, $\mathcal{L}_d$ is edge-preserving and works only on small areas, avoiding over-smoothing in most cases.

This prior is reasonable under our setting since most of the reflectors in the real world are planar surfaces, which exhibit smooth depth changes in the reflected areas. 
Local depth smoothness priors are widely adopted in traditional stereo matching~\cite{sinha2009piecewise,sinha2012image,zhang2014rigid}, but rarely explored in the training of neural representations. To apply this regularization during training, we sample patches instead of pixels.

\paragraph{Bidirectional depth consistency.}

\begin{figure}[tb]
  \centering
   \includegraphics[width=1.0\linewidth]{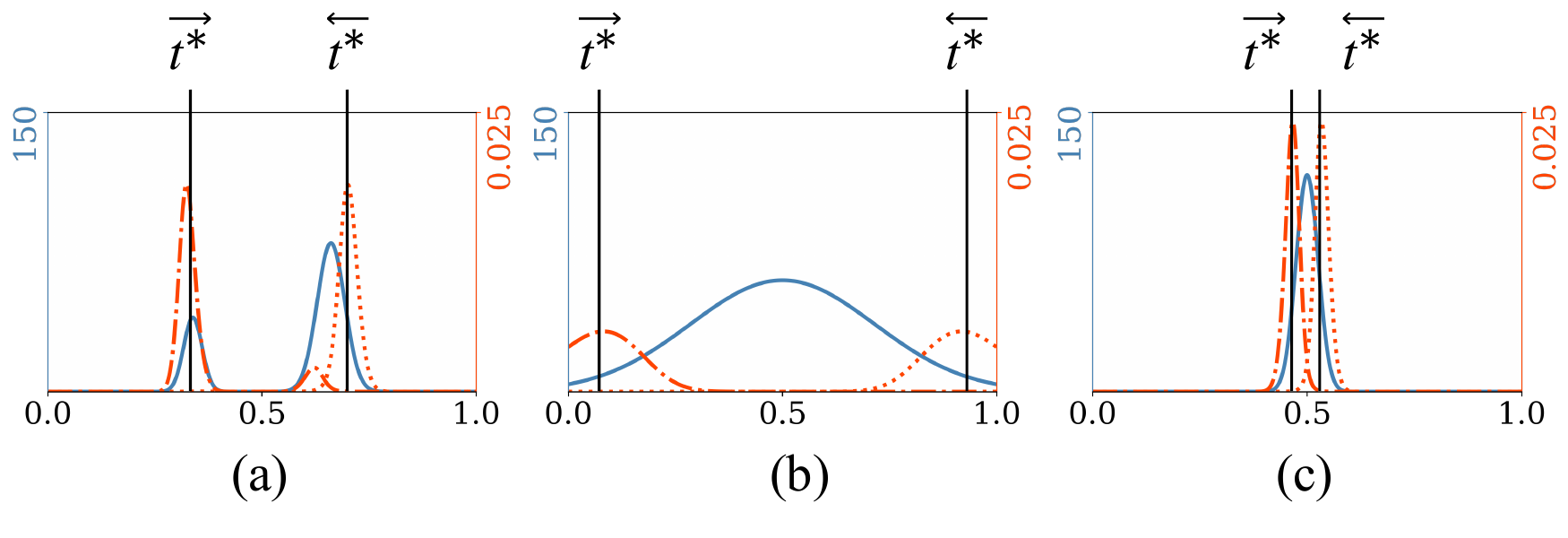}
   \caption{Illustrations for bidirectional depth consistency. 
   Here we list three common density distributions (solid blue curve)
   along a ray in a neural radiance field. 
   For each graph, the horizontal axis represents point locations along a ray. 
   The dashdot and dotted orange curve are point weights (\cref{eqn:point-weight,eqn:backward-depth}) at each location for calculating the forward and backward depth (\cref{eqn:depth,eqn:backward-depth}) respectively. 
   The computed depth values are marked with solid black lines. 
   Only in case (c) we have a low BDC value, which corresponds to the expected shell-like geometry.
   }
   \label{fig:bdc}
   \vspace{-0.5cm}
\end{figure}

The depth smoothness prior encourages correct geometry for the transmitted part. 
We now put constraints on the reflected part by assuming it has a simple geometry, where each ray only hits a single opaque surface. This assumption can help since we can only see the reflected objects from limited viewing angles so that they are, to some extend, similar to a textured \emph{solid shell}.
To describe this kind of simplicity in mathematical form, we propose a novel bidirectional depth consistency prior to encourage the aforementioned characteristic. 
We first define a new \emph{backward depth} $\overleftarrow{t}^*$ along ray $\mathbf{r}$ as the expected termination point seeing the volume from the opposite direction of $\mathbf{r}$:
\begin{equation}\label{eqn:backward-depth}
\overleftarrow{t}^*(\mathbf{r};\sigma)=\sum_{k}\overleftarrow{\omega}_{i}t_{i}=\sum_{k}\overleftarrow{T}_i(\sigma)(1-\text{exp}(-\sigma_i\delta_i))t_i
\end{equation}
where $\overleftarrow{T}_i(\sigma)=\text{exp}(-\sum_{j>i}\sigma_j\delta_j)$. Denote $\overrightarrow{t}^*$ as previous $t^*$, then our proposed bidirectional depth consistency (BDC) is defined as:
\begin{equation}
\mathcal{L}_{bdc}=\sum_{\mathbf{r}}||\overrightarrow{t}^*(\mathbf{r};\sigma^r)-\overleftarrow{t}^*(\mathbf{r};\sigma^r)||_1
\end{equation}
This regularization poses restriction on the density distribution along a ray, forcing it to be uni-modal and have small variance. 
The idea is demonstrated in \cref{fig:bdc}. 
Here we consider three representative types of density distribution along the ray in a neural radiance field. 
If the ray hits multiple surfaces as in (a) or marches in a ``foggy" area without a clear surface as in (b), the forward and backward depth is not consistent as marked in the figure. 
The two depth values become close only if the density distribution is uni-modal and has small variance as in (c). 
Intuitively, this requires the reflected component to have a ``shell"-like geometry. Note that $\mathcal{L}_{bdc}=0$ could also correspond to an empty ray. We claim that this would not happen because the transmitted component alone cannot faithfully reproduce the input images due to the existence of the depth smoothness prior. Some concurrent works~\cite{rematas2021urban,roessle2021dense} also address the density distribution problem and propose regularizations to concentrate the densities at some \textbf{known depth}, while BDC achieves this with \textbf{unknown depth}.

\subsection{Warm-up Training}\label{sec:training}

The overall loss used to optimize {\sysname} is the weighted combination of photometric loss and the proposed geometric priors:
\begin{equation}
\mathcal{L}=\mathcal{L}_{pm}+\lambda_{d}\mathcal{L}_{d}+\lambda_{bdc}\mathcal{L}_{bdc}
\end{equation}
The magnitude of the geometric priors can greatly influence the decomposition results, requiring careful tuning of the weighting factors which can hardly be shared across scenes.
It is difficult to get a balance between the photometric loss and the geometric constraints especially at the early stage of training.
If the geometric constraints dominate, the model will stuck at a bad local minima by using only one of the components to explain the whole scene while leaving the other empty. 
On the other hand, if the photometric loss dominates, insufficient geometric regularizations will lead to sub-optimal decomposition results.

Based on the above observations, we propose a universal training strategy to avoid the need for tuning hyper-parameters for each scene which effectively stabilizes the training process. We take inspirations from learning rate warm-up and propose to ``warm-up" the geometric constraints. $\lambda_{d}$ and $\lambda_{bdc}$ are initialized with small values, and are first increased then decreased as the training progresses. We also mask out the input viewing direction at early stage of training to avoid ``leakage" of the reflected component into the transmitted one. We refer to the supplementary material for more details.

\subsection{Interactive Setting}\label{sec:interactive}

The proposed geometric priors and training strategy work well when the transmitted geometry can be correctly estimated from the images. 
However, there exist more challenging cases, such as texture-less reflectors like mirrors, where the unsupervised decomposition often fails. 
Thanks to the additive formulation in \cref{eqn:formulation-image,eqn:formulation}, we can utilize extra information such as manually labeled reflection fraction maps. 

\begin{figure*}[t]
  \centering
   \includegraphics[width=1.0\linewidth]{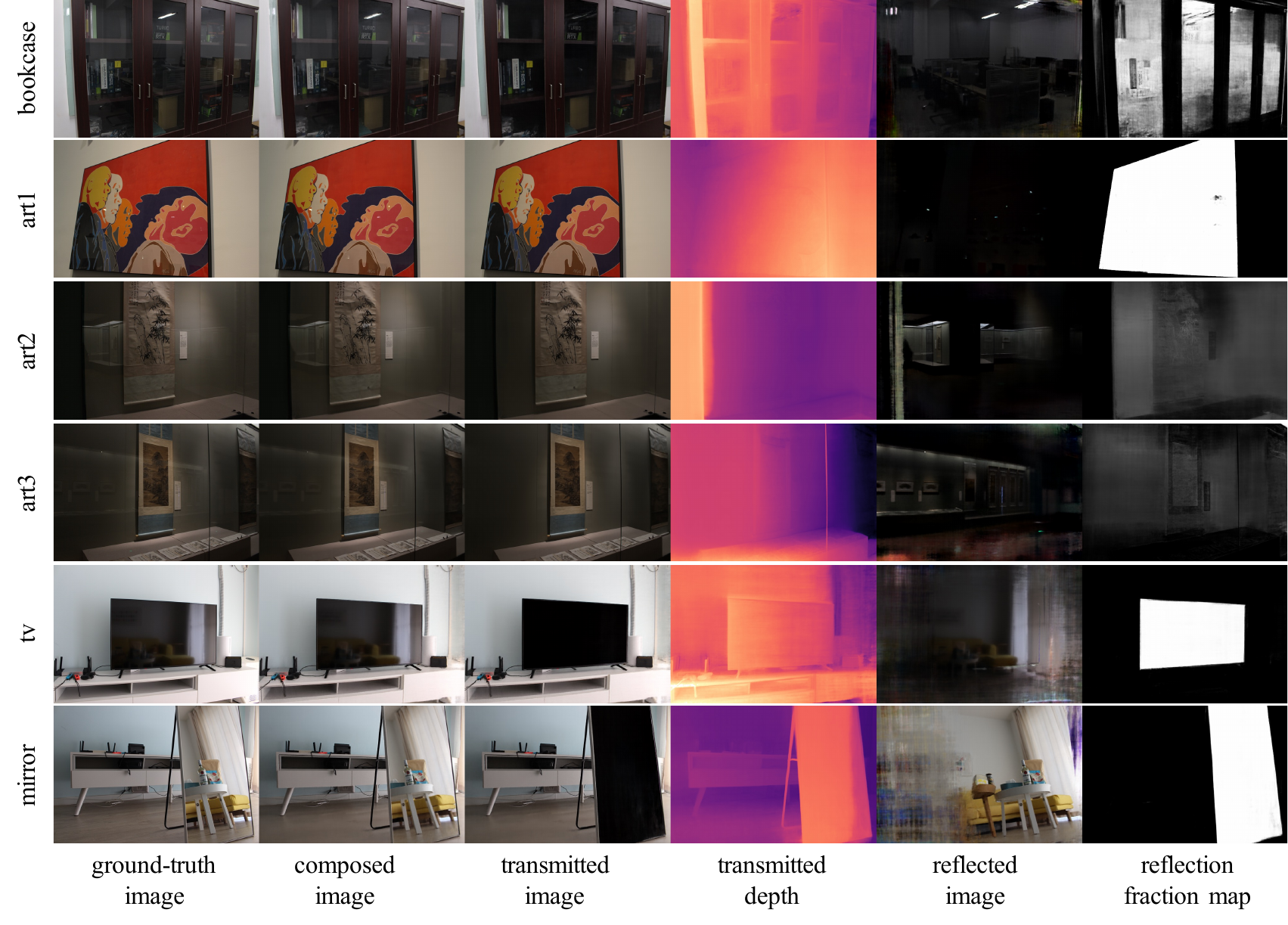}
   \caption{Decomposition results of {\sysname} on RFFR scenes. 
   The decomposed images and depth estimations are highly consistent with human perception.}
   \label{fig:decomposition}
  \vspace{-0.5cm}
\end{figure*}

We exploit an interactive setting where the user provides binary masks of a small number of training images, with 1 and 0 denoting reflective and non-reflective regions respectively. 
L1 loss is used to encourage consistency between the predicted reflection fraction map $\beta$ and the user-provided masks:
\begin{equation}
\mathcal{L}_{\beta}=\sum_{p}||\widehat{\beta}(p)-\beta(p)||_1
\end{equation}
where $\widehat{\beta}(p)=\beta(\mathbf{r}(p);\sigma^t,\alpha)$ is the estimation of the reflection fraction value, and $\beta(p)$ is the value of the user-provided binary mask at pixel $p$. 
With the help of this extra supervision, we are able to successfully isolate the reflected component in several challenging scenarios as demonstrated in \cref{fig:decomposition}. These scenarios are hard to deal with using existing novel view synthesis techniques.

\section{Experiments}

In this section, we first show the scene decomposition results in \cref{sec:exp-decomp}. 
Then, we compare with NeRF~\cite{mildenhall2020nerf} and its {\naive} variant to show that {\sysname} achieves comparable visual quality with significantly better depth estimation results in \cref{sec:exp-compare}.
We validate some of our design choices in \cref{sec:exp-ablation}, and finally provide applications for reflection removal and scene editing in \cref{sec:exp-app}.

\topic{Baselines.}
We compare {\sysname} with NeRF and a variant of NeRF which we call NeRF-D. 
NeRF-D applies our proposed depth smoothness constraint on the original NeRF, which is expected to model the reflected image solely by the surface point itself rather than all the points behind it along this viewing direction.

\topic{Datasets.} 
Due to the lack of real world datasets with strong reflections for the view synthesis task, we introduce RFFR (Real Forward-Facing with Reflections) dataset, containing 6 captured forward-facing scenes with strong reflection effects caused by glass and mirrors. 
We split the images into training and test set, and report the qualitative and quantitative results on test images. 

\begin{figure}[t]
  \centering
   \includegraphics[width=1.0\linewidth]{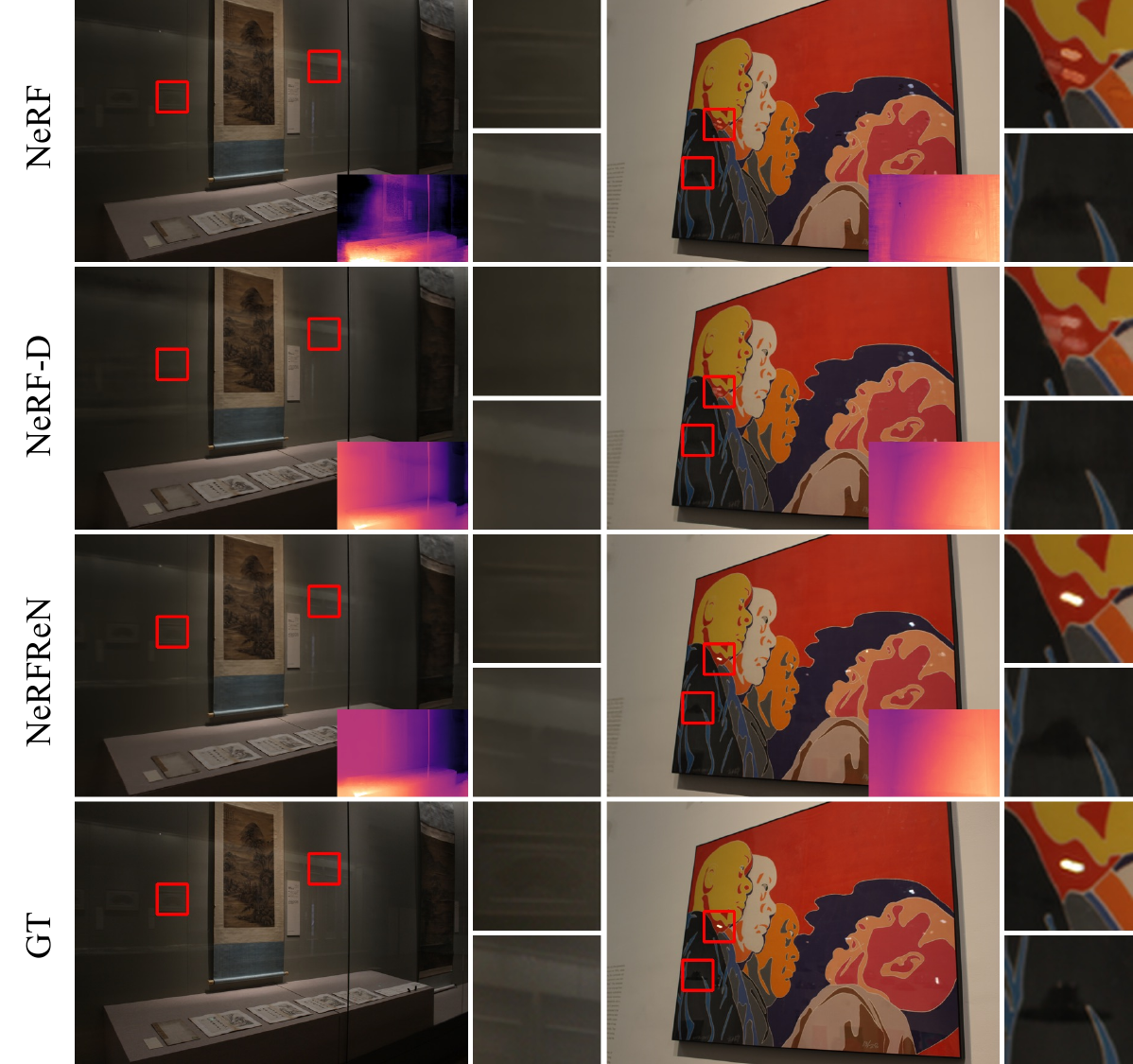}
   \caption{Visual comparisons between NeRF, NeRF-D and {\sysname}. {\sysname} achieves comparable or better novel view synthesis quality with physically correct depth estimation. Zoom in for more details.}
   \label{fig:compare}
  \vspace{-0.5cm}
\end{figure}

\begin{figure}[t]
  \centering
   \includegraphics[width=1.0\linewidth]{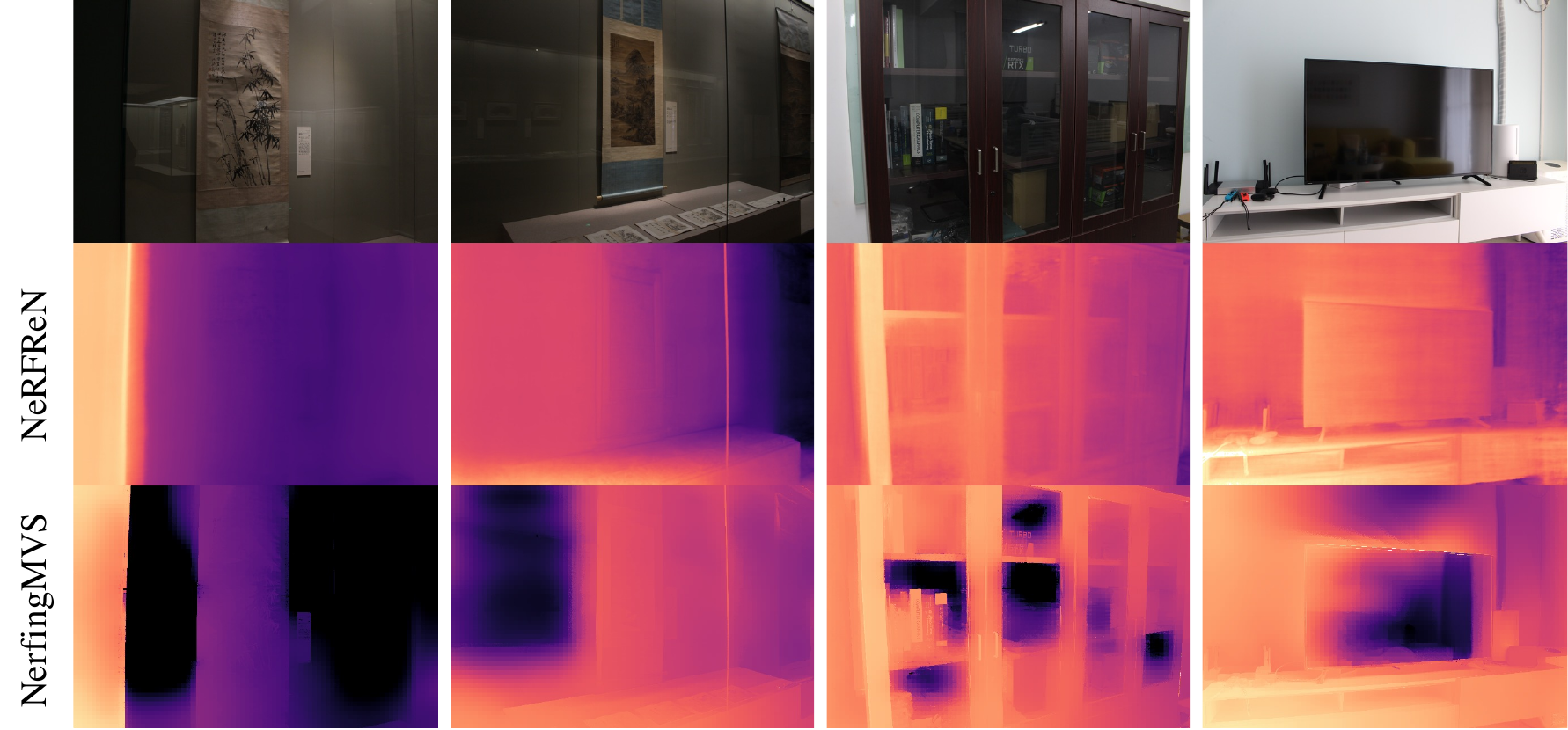}
   \caption{Qualitative comparisons with NerfingMVS~\cite{wei2021nerfingmvs} on depth estimation. NerfingMVS fails in such scenes with severe reflections, behaving similar to NeRF.}
   \label{fig:compare-depth}
  \vspace{-0.5cm}
\end{figure}

\subsection{Decomposition}\label{sec:exp-decomp}

In \cref{fig:decomposition} we show how {\sysname} decomposes the scene and performs view synthesis on our RFFR scenes. 
The transmitted image, reflected image, and reflection fraction map are rendered from the predicted transmitted and reflected neural radiance fields. 
The composed image is generated by combining the transmitted image and the reflected image weighted by the reflection fraction map according to \cref{eqn:formulation-image}. 
The depth values are computed by \cref{eqn:depth}. For \textit{tv} and \textit{mirror}, 1 and 4 manually-annotated reflection masks are utilized respectively as extra information, as described in \cref{sec:interactive}. For rest of the scenes, the decomposition is inferred from the training images alone.

We can see from \cref{fig:decomposition} that {\sysname} can perform transmission-reflection decomposition that is consistent with human perception while producing high-quality view synthesis results. 
The learned reflection fraction map roughly indicates the surface material, with large values (white) for reflectors and small values (black) for non-reflectors. 
Note that we only intend to model the stable virtual image by the reflected field, and leave the low-frequency highlights to the view-dependency in the transmitted field. 
For example, there are highlights on the frames of the bookcase, where the reflection fraction is almost zero.

Accurately estimating the location of transparent surfaces like glass is difficult for lack of traceable image features. In such cases like \textit{art2} and \textit{art3}, NeRFReN treats the opaque surface behind as the reflective surface, which is not strictly correct but does not affect view synthesis quality due to the image-space composition.

\begin{table}
  \centering
  \footnotesize
  \renewcommand\tabcolsep{5pt}
  \begin{tabular}{@{}l|cccccc@{}}
    \toprule
    \multicolumn{1}{c}{} & \multicolumn{6}{c}{PSNR$\uparrow$} \\
    Method & art1 & art2 & art3 & bookcase & tv & mirror \\
    \midrule
    NeRF & 36.48 & \textbf{42.63} & \textbf{41.18} & 29.50 & \textbf{33.47} & 32.22 \\
    NeRF-D & \underline{36.49} & 41.58 & 39.80 & \underline{29.78} & 33.02 & \underline{32.33} \\
    NeRFReN & \textbf{39.49} & \underline{42.33} & \underline{41.03} & \textbf{30.18} & \underline{33.24} & \textbf{33.30}\\
    \midrule
    \multicolumn{1}{c}{} & \multicolumn{6}{c}{SSIM$\uparrow$} \\
    Method & art1 & art2 & art3 & bookcase & tv & mirror \\
    \midrule
    NeRF & \underline{0.975} & \textbf{0.974} & \textbf{0.973} & \underline{0.883} & \textbf{0.960} & \underline{0.944} \\
    NeRF-D & 0.971 & 0.968 & 0.965 & 0.876 & 0.954 & 0.939\\
    NeRFReN & \textbf{0.978} & \underline{0.972} & \underline{0.970} & \textbf{0.890} & \underline{0.955} & \textbf{0.947} \\
    \midrule
    \multicolumn{1}{c}{} & \multicolumn{6}{c}{LPIPS$\downarrow$} \\
    Method & art1 & art2 & art3 & bookcase & tv & mirror \\
    \midrule
    NeRF & \underline{0.029} & \textbf{0.059} & \textbf{0.053} & \textbf{0.078} & \textbf{0.054} & \textbf{0.066} \\
    NeRF-D & \textbf{0.027} & 0.070 & 0.117 & 0.136 & 0.063 & 0.072 \\
    NeRFReN & \underline{0.029} & \underline{0.067} & \underline{0.072} & \underline{0.100} & \underline{0.061} & \underline{0.071} \\
    \bottomrule
  \end{tabular}
  \caption{View synthesis results of NeRF, NeRF-D and {\sysname} on RFFR scenes in PSNR, SSIM~\cite{wang2003multiscale} and LPIPS~\cite{zhang2018unreasonable}. {\sysname} performs comparable with NeRF and better than NeRF-D.}
  \label{tab:compare}
  \vspace{-0.5cm}
\end{table}

\subsection{Comparisons}\label{sec:exp-compare}
We compare {\sysname} with NeRF and NeRF+D on the test set of our RFFR dataset under three metrics: PSNR, SSIM~\cite{wang2003multiscale} and LPIPS~\cite{zhang2018unreasonable} (see \cref{tab:compare}). 
We also provide qualitative comparisons in \cref{fig:compare}.

Quantitatively, {\sysname} significantly outperforms NeRF on \textit{art1} and \textit{mirror} in terms of PSNR. While performing slightly better on rest of the scenes, NeRF produces inaccurate depth estimation due to the mixed geometry of the transmitted and reflected components as demonstrated in \cref{sec:nerf}, which is shown on the bottom left corner of the NeRF rendering results in \cref{fig:compare}. {\sysname} models the scene with separate neural radiance fields, and provides physically correct depth map by taking the depth estimation of the transmitted field. Also, NeRF generates sub-optimal renderings when the multi-view consistency is violated while {\sysname} does not suffer from such restrictions, as shown in the \textit{mirror} case in \cref{fig:teaser}(b). NeRF-D produces more correct depth estimation than NeRF by adopting the depth smoothness regularization, and condition the surface color on viewing direction. However, it synthesizes blurry results due to the insufficient representation power of NeRF's view-dependency in modeling complex reflections. The quantitative results conform with this claim that NeRF-D consistently underperforms NeRF and {\sysname}.

To demonstrate the depth estimation quality of {\sysname}, we also qualitatively compare with NerfingMVS~\cite{wei2021nerfingmvs} on RFFR scenes. Results are shown in \cref{fig:compare-depth}. NerfingMVS utilizes monocular depth estimation priors and post-filtering to achieve promising results on ScanNet scenes, but struggles on scenes with complex reflections. This is because NerfingMVS initializes the depth maps based on Multi-View-Stereo output, which cannot give faithful depth estimations in such cases.

To conclude, {\sysname} achieves visually pleasing view synthesis results and physically correct depth estimation results at the same time. The formulation of {\sysname} naturally relax the multi-view consistency of NeRF by combining the transmitted and reflected part in the image domain, which greatly facilitate novel view synthesis in challenging scenes like those with mirrors.

\subsection{Ablation Studies}\label{sec:exp-ablation}

We discuss the necessity of several key design choices, and demonstrate their importance to getting faithful decompositions and correct view synthesis results.

\begin{figure}[t]
  \centering
   \includegraphics[width=1.0\linewidth]{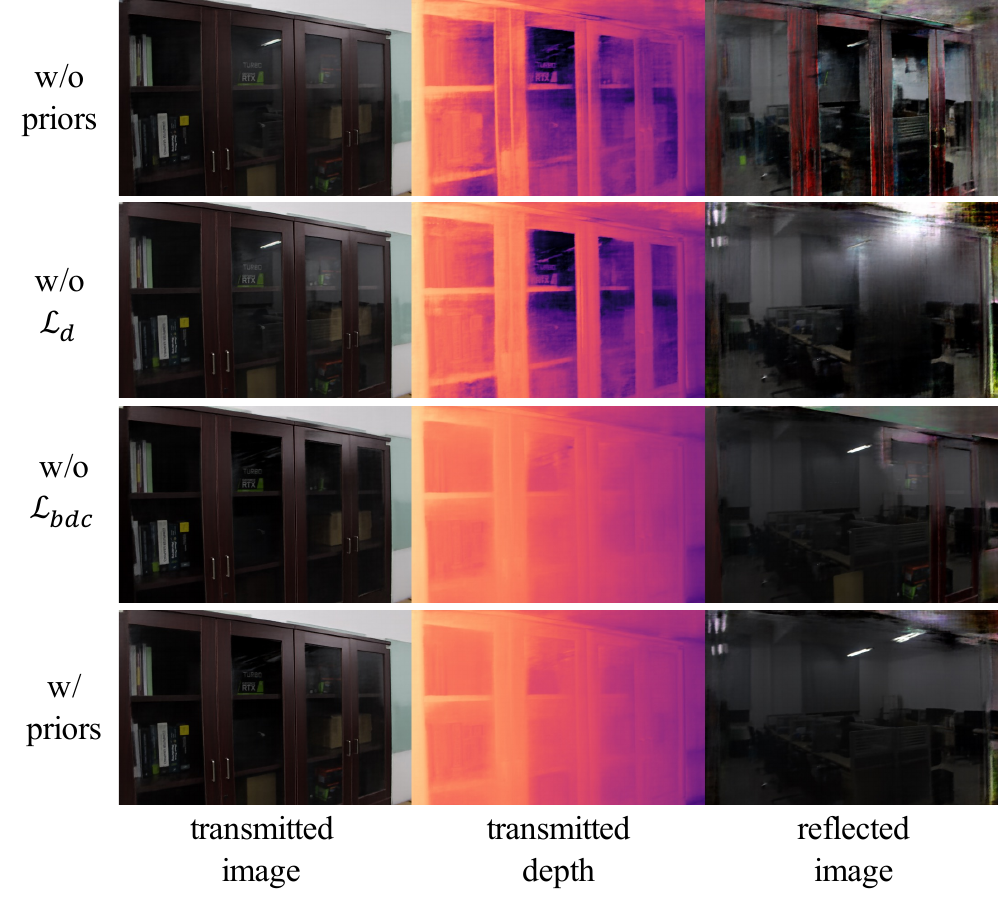}
   \caption{Effectiveness of the proposed geometric priors. Both priors are essential to get physically sound decompositions.}
   \label{fig:ablation-prior}
  \vspace{-0.1cm}
\end{figure}

\topic{Geometric priors.}
We train {\sysname} without any geometric priors and with only one of the priors, to see the impact of these priors on the decomposition quality. 
The results are shown in \cref{fig:ablation-prior}. 
Without any priors (row 1), we find the optimization process to be highly under-controlled, largely depending on the network initialization. 
In this case, the decomposition results are not reasonable since the transmitted and reflected part are mixed together. 
Without the depth smoothness prior (row 2), there is not enough constraint on the transmitted geometry, letting it contain reflected objects and producing wrong depth. 
Without the bidirectional depth consistency prior (row 3), there will be redundant contents left in the reflected component (the bookcase frame). The BDC loss encourages uni-modal peak distribution for densities along a ray, which effectively removes occlusions to get a “clean” reflected image.
We claim that both priors are essential to getting reasonable solutions for this under-constrained problem (row 4).

\begin{figure}[t]
  \centering
   \includegraphics[width=1.0\linewidth]{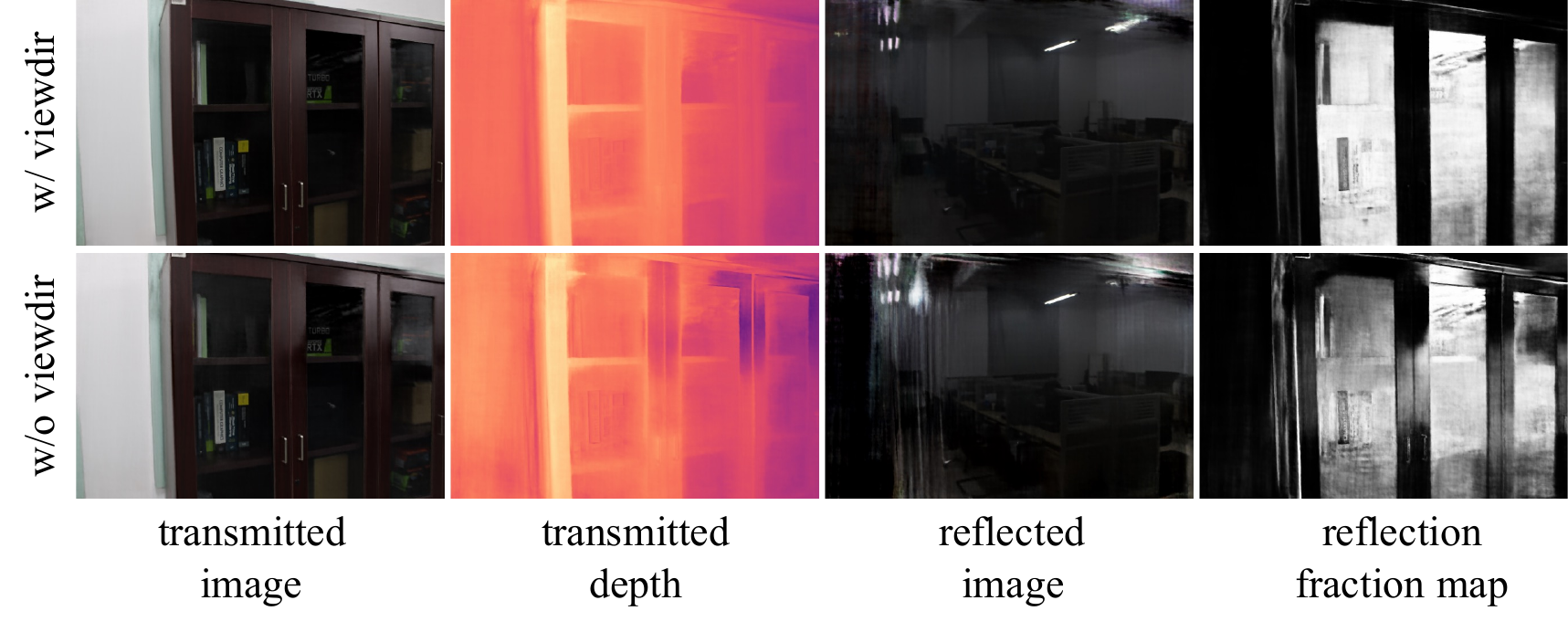}
   \caption{The effectiveness of conditioning transmitted radiance on viewing direction. 
   Without this dependency, low-frequency reflections like highlights are modeled with foggy contents in the reflected component, leading to inaccurate depth estimation.}
   \label{fig:ablation-viewdir}
  \vspace{-0.5cm}
\end{figure}

\topic{View-dependency.}
{\sysname} conditions the color of the transmitted component on viewing direction to model low-frequency view-dependent effects. 
We remove this dependency on the \textit{bookcase} scene and show the results in \cref{fig:ablation-viewdir}. 
Without the view-dependency, highlights on the bookcase frame are modeled with the combined effect of the reflected image and the reflection fraction map. 
This results in inaccurate depth estimation (dark regions on the predicted depth map), and redundancies in the reflected component which do not belong to the virtual geometry.

\begin{figure}
  \centering
   \includegraphics[width=1.0\linewidth]{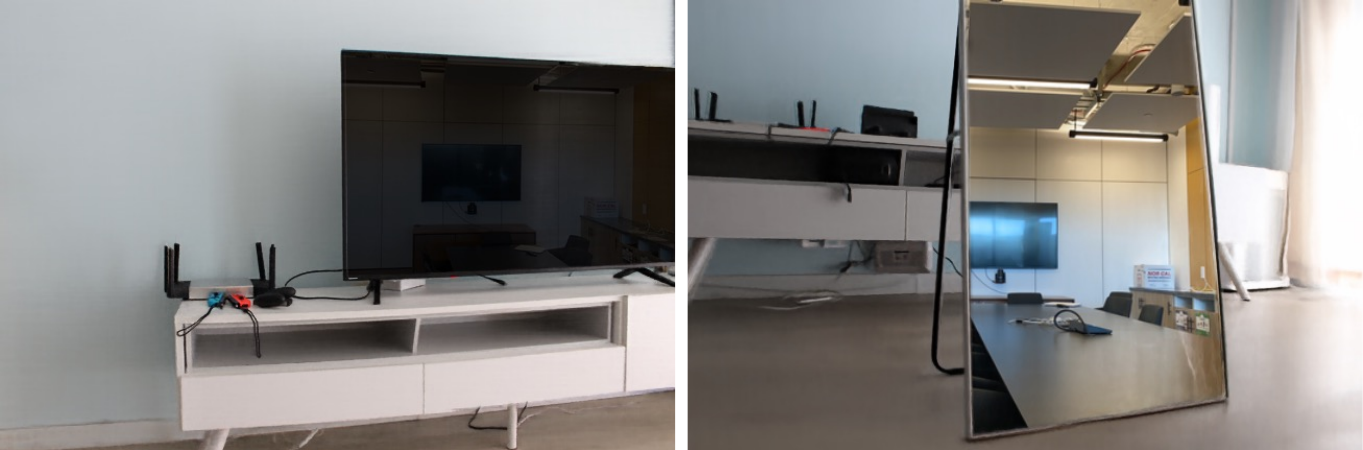}
   \caption{Examples of replacing the reflected image (with the \textit{room} scene in~\cite{mildenhall2020nerf}). This could be utilized to synthesize self-reflections of the user for higher realism.}
   \label{fig:app-sub}
\end{figure}

\subsection{Applications}\label{sec:exp-app}

Enabling user editing for neural scene representations has been attracting attention in the community~\cite{guo2020object,liu2021editing,yang2021learning,srinivasan2021nerv,boss2021nerd,zhang2021nerfactor}. Existing techniques either achieve material editing or relighting by reflectance decomposition~\cite{srinivasan2021nerv,boss2021nerd,zhang2021nerfactor} or model the scene at object level for manipulation~\cite{guo2020object,liu2021editing,yang2021learning}. Benefit from the formulation in \cref{eqn:formulation-image}, {\sysname} enables scene editing in the image domain while maintaining view consistency, which is more intuitive and makes it possible to work with other scene representations.

\topic{Reflection removal.}
Our formulation naturally supports the application for reflection removal by taking only the transmitted image. See column 3 of \cref{fig:decomposition}.

\topic{Reflection substitution.}
We achieve reflection substitution by replacing the reflection image $I_r$ by images coming from other neural radiance fields, or even from other scene representations like mesh. \cref{fig:app-sub} shows two examples of replacing the reflections with images rendered from another NeRF model trained on the \textit{room} scene~\cite{mildenhall2020nerf}. This could be further promoted to synthesize self-reflections of the user to provide even more immersive experiences. We hope the image-based formulation of {\sysname} can inspire future research about combining multiple scene representations.

\section{Limitations and Conclusion}
Some major limitations of our method include modeling curved reflective surfaces and multiple non-coplanar reflective surfaces. Also, we do not consider the view-dependent effects of reflected objects, and ignore the view-dependency of reflection fraction values caused by Fresnel effect.

To conclude, we propose {\sysname} to enhance NeRF in scenes with reflections. 
We decompose the scene into transmitted and reflected components, model them with separate neural radiance fields, and synthesize novel views by weighted combination. 
We exploit geometric priors and special training strategies to encourage proper decomposition, where user-provided reflection masks can be utilized in challenging cases like mirrors. 
Our method enables high-quality view synthesis along with physically sound decomposition results.
Finally, potential applications including depth estimation, reflection removal and reflection substitution are investigated to encourage further research about scene understanding and neural editing.

\paragraph{Acknowledgements.} This work was supported by the National Natural Science Foundation of China (Project Number 62132012), Research Grant of Beijing Higher Institution Engineering Research Center, and Tsinghua–Tencent Joint Laboratory for Internet Innovation Technology.

{\small
\bibliographystyle{ieee_fullname}
\bibliography{egbib}

\begin{thebibliography}{10}\itemsep=-1pt

\bibitem{aliev2020neural}
Kara-Ali Aliev, Artem Sevastopolsky, Maria Kolos, Dmitry Ulyanov, and Victor
  Lempitsky.
\newblock Neural point-based graphics.
\newblock In {\em Computer Vision--ECCV 2020: 16th European Conference,
  Glasgow, UK, August 23--28, 2020, Proceedings, Part XXII 16}, pages 696--712.
  Springer, 2020.

\bibitem{boss2021nerd}
Mark Boss, Raphael Braun, Varun Jampani, Jonathan~T Barron, Ce Liu, and Hendrik
  Lensch.
\newblock Nerd: Neural reflectance decomposition from image collections.
\newblock In {\em Proceedings of the IEEE/CVF International Conference on
  Computer Vision}, pages 12684--12694, 2021.

\bibitem{buehler2001unstructured}
Chris Buehler, Michael Bosse, Leonard McMillan, Steven Gortler, and Michael
  Cohen.
\newblock Unstructured lumigraph rendering.
\newblock In {\em Proceedings of the 28th annual conference on Computer
  graphics and interactive techniques}, pages 425--432, 2001.

\bibitem{chen2019learning}
Zhiqin Chen and Hao Zhang.
\newblock Learning implicit fields for generative shape modeling.
\newblock In {\em Proceedings of the IEEE/CVF Conference on Computer Vision and
  Pattern Recognition}, pages 5939--5948, 2019.

\bibitem{flynn2019deepview}
John Flynn, Michael Broxton, Paul Debevec, Matthew DuVall, Graham Fyffe, Ryan
  Overbeck, Noah Snavely, and Richard Tucker.
\newblock Deepview: View synthesis with learned gradient descent.
\newblock In {\em Proceedings of the IEEE/CVF Conference on Computer Vision and
  Pattern Recognition}, pages 2367--2376, 2019.

\bibitem{guo2020object}
Michelle Guo, Alireza Fathi, Jiajun Wu, and Thomas Funkhouser.
\newblock Object-centric neural scene rendering.
\newblock {\em arXiv preprint arXiv:2012.08503}, 2020.

\bibitem{hedman2016scalable}
Peter Hedman, Tobias Ritschel, George Drettakis, and Gabriel Brostow.
\newblock Scalable inside-out image-based rendering.
\newblock {\em ACM Transactions on Graphics (TOG)}, 35(6):1--11, 2016.

\bibitem{kingma2014adam}
Diederik~P Kingma and Jimmy Ba.
\newblock Adam: A method for stochastic optimization.
\newblock {\em arXiv preprint arXiv:1412.6980}, 2014.

\bibitem{liu2021editing}
Steven Liu, Xiuming Zhang, Zhoutong Zhang, Richard Zhang, Jun-Yan Zhu, and
  Bryan Russell.
\newblock Editing conditional radiance fields.
\newblock {\em arXiv preprint arXiv:2105.06466}, 2021.

\bibitem{lombardi2019neural}
Stephen Lombardi, Tomas Simon, Jason Saragih, Gabriel Schwartz, Andreas
  Lehrmann, and Yaser Sheikh.
\newblock Neural volumes: Learning dynamic renderable volumes from images.
\newblock {\em arXiv preprint arXiv:1906.07751}, 2019.

\bibitem{max1995optical}
Nelson Max.
\newblock Optical models for direct volume rendering.
\newblock {\em IEEE Transactions on Visualization and Computer Graphics},
  1(2):99--108, 1995.

\bibitem{mescheder2019occupancy}
Lars Mescheder, Michael Oechsle, Michael Niemeyer, Sebastian Nowozin, and
  Andreas Geiger.
\newblock Occupancy networks: Learning 3d reconstruction in function space.
\newblock In {\em Proceedings of the IEEE/CVF Conference on Computer Vision and
  Pattern Recognition}, pages 4460--4470, 2019.

\bibitem{mildenhall2020nerf}
Ben Mildenhall, Pratul~P Srinivasan, Matthew Tancik, Jonathan~T Barron, Ravi
  Ramamoorthi, and Ren Ng.
\newblock Nerf: Representing scenes as neural radiance fields for view
  synthesis.
\newblock In {\em European conference on computer vision}, pages 405--421.
  Springer, 2020.

\bibitem{park2019deepsdf}
Jeong~Joon Park, Peter Florence, Julian Straub, Richard Newcombe, and Steven
  Lovegrove.
\newblock Deepsdf: Learning continuous signed distance functions for shape
  representation.
\newblock In {\em Proceedings of the IEEE/CVF Conference on Computer Vision and
  Pattern Recognition}, pages 165--174, 2019.

\bibitem{rematas2021urban}
Konstantinos Rematas, Andrew Liu, Pratul~P Srinivasan, Jonathan~T Barron,
  Andrea Tagliasacchi, Thomas Funkhouser, and Vittorio Ferrari.
\newblock Urban radiance fields.
\newblock {\em arXiv preprint arXiv:2111.14643}, 2021.

\bibitem{riegler2020free}
Gernot Riegler and Vladlen Koltun.
\newblock Free view synthesis.
\newblock In {\em European Conference on Computer Vision}, pages 623--640.
  Springer, 2020.

\bibitem{riegler2021stable}
Gernot Riegler and Vladlen Koltun.
\newblock Stable view synthesis.
\newblock In {\em Proceedings of the IEEE/CVF Conference on Computer Vision and
  Pattern Recognition}, pages 12216--12225, 2021.

\bibitem{roessle2021dense}
Barbara Roessle, Jonathan~T Barron, Ben Mildenhall, Pratul~P Srinivasan, and
  Matthias Nie{\ss}ner.
\newblock Dense depth priors for neural radiance fields from sparse input
  views.
\newblock {\em arXiv preprint arXiv:2112.03288}, 2021.

\bibitem{schoenberger2016sfm}
Johannes~Lutz Sch\"{o}nberger and Jan-Michael Frahm.
\newblock Structure-from-motion revisited.
\newblock In {\em Conference on Computer Vision and Pattern Recognition
  (CVPR)}, 2016.

\bibitem{schoenberger2016mvs}
Johannes~Lutz Sch\"{o}nberger, Enliang Zheng, Marc Pollefeys, and Jan-Michael
  Frahm.
\newblock Pixelwise view selection for unstructured multi-view stereo.
\newblock In {\em European Conference on Computer Vision (ECCV)}, 2016.

\bibitem{sinha2012image}
Sudipta~N Sinha, Johannes Kopf, Michael Goesele, Daniel Scharstein, and Richard
  Szeliski.
\newblock Image-based rendering for scenes with reflections.
\newblock {\em ACM Transactions on Graphics (TOG)}, 31(4):1--10, 2012.

\bibitem{sinha2009piecewise}
Sudipta~N Sinha, Drew Steedly, and Richard Szeliski.
\newblock Piecewise planar stereo for image-based rendering.
\newblock In {\em 2009 IEEE 12th International Conference on Computer Vision},
  pages 1881--1888. IEEE, 2009.

\bibitem{sitzmann2019deepvoxels}
Vincent Sitzmann, Justus Thies, Felix Heide, Matthias Nie{\ss}ner, Gordon
  Wetzstein, and Michael Zollhofer.
\newblock Deepvoxels: Learning persistent 3d feature embeddings.
\newblock In {\em Proceedings of the IEEE/CVF Conference on Computer Vision and
  Pattern Recognition}, pages 2437--2446, 2019.

\bibitem{sitzmann2019scene}
Vincent Sitzmann, Michael Zollh{\"o}fer, and Gordon Wetzstein.
\newblock Scene representation networks: Continuous 3d-structure-aware neural
  scene representations.
\newblock {\em arXiv preprint arXiv:1906.01618}, 2019.

\bibitem{snavely2006photo}
Noah Snavely, Steven~M Seitz, and Richard Szeliski.
\newblock Photo tourism: exploring photo collections in 3d.
\newblock In {\em ACM siggraph 2006 papers}, pages 835--846. 2006.

\bibitem{srinivasan2021nerv}
Pratul~P Srinivasan, Boyang Deng, Xiuming Zhang, Matthew Tancik, Ben
  Mildenhall, and Jonathan~T Barron.
\newblock Nerv: Neural reflectance and visibility fields for relighting and
  view synthesis.
\newblock In {\em Proceedings of the IEEE/CVF Conference on Computer Vision and
  Pattern Recognition}, pages 7495--7504, 2021.

\bibitem{thies2019deferred}
Justus Thies, Michael Zollh{\"o}fer, and Matthias Nie{\ss}ner.
\newblock Deferred neural rendering: Image synthesis using neural textures.
\newblock {\em ACM Transactions on Graphics (TOG)}, 38(4):1--12, 2019.

\bibitem{wang2003multiscale}
Zhou Wang, Eero~P Simoncelli, and Alan~C Bovik.
\newblock Multiscale structural similarity for image quality assessment.
\newblock In {\em The Thrity-Seventh Asilomar Conference on Signals, Systems \&
  Computers, 2003}, volume~2, pages 1398--1402. Ieee, 2003.

\bibitem{wei2021nerfingmvs}
Yi Wei, Shaohui Liu, Yongming Rao, Wang Zhao, Jiwen Lu, and Jie Zhou.
\newblock Nerfingmvs: Guided optimization of neural radiance fields for indoor
  multi-view stereo.
\newblock In {\em Proceedings of the IEEE/CVF International Conference on
  Computer Vision}, pages 5610--5619, 2021.

\bibitem{wizadwongsa2021nex}
Suttisak Wizadwongsa, Pakkapon Phongthawee, Jiraphon Yenphraphai, and Supasorn
  Suwajanakorn.
\newblock Nex: Real-time view synthesis with neural basis expansion.
\newblock In {\em Proceedings of the IEEE/CVF Conference on Computer Vision and
  Pattern Recognition}, pages 8534--8543, 2021.

\bibitem{xu2021scalable}
Jiamin Xu, Xiuchao Wu, Zihan Zhu, Qixing Huang, Yin Yang, Hujun Bao, and Weiwei
  Xu.
\newblock Scalable image-based indoor scene rendering with reflections.
\newblock {\em ACM Transactions on Graphics (TOG)}, 40(4):1--14, 2021.

\bibitem{yang2021learning}
Bangbang Yang, Yinda Zhang, Yinghao Xu, Yijin Li, Han Zhou, Hujun Bao, Guofeng
  Zhang, and Zhaopeng Cui.
\newblock Learning object-compositional neural radiance field for editable
  scene rendering.
\newblock In {\em Proceedings of the IEEE/CVF International Conference on
  Computer Vision}, pages 13779--13788, 2021.

\bibitem{zhang2014rigid}
Chi Zhang, Zhiwei Li, Rui Cai, Hongyang Chao, and Yong Rui.
\newblock As-rigid-as-possible stereo under second order smoothness priors.
\newblock In {\em European Conference on Computer Vision}, pages 112--126.
  Springer, 2014.

\bibitem{zhang2018unreasonable}
Richard Zhang, Phillip Isola, Alexei~A Efros, Eli Shechtman, and Oliver Wang.
\newblock The unreasonable effectiveness of deep features as a perceptual
  metric.
\newblock In {\em Proceedings of the IEEE conference on computer vision and
  pattern recognition}, pages 586--595, 2018.

\bibitem{zhang2021nerfactor}
Xiuming Zhang, Pratul~P Srinivasan, Boyang Deng, Paul Debevec, William~T
  Freeman, and Jonathan~T Barron.
\newblock Nerfactor: Neural factorization of shape and reflectance under an
  unknown illumination.
\newblock {\em arXiv preprint arXiv:2106.01970}, 2021.

\end{thebibliography}
}

\clearpage

\setcounter{section}{0}
\renewcommand{\thesection}{A.\arabic{section}}
\setcounter{equation}{0}
\renewcommand{\theequation}{A.\arabic{equation}}
\setcounter{table}{0}
\renewcommand{\thetable}{A\arabic{table}}
\setcounter{figure}{0}
\renewcommand{\thefigure}{A\arabic{figure}}

\section*{Supplementary Material}

In this supplementary material, we provide additional details for the network and training procedure (\cref{sec:network-and-training}), further discussions on limitations along with an example of failure case (\cref{sec:failure-case-example}), results and comparisons on LLFF Dataset~\cite{mildenhall2020nerf} (\cref{sec:llff}), and more ablation studies with qualitative demonstrations (\cref{sec:ablation-studies}). We also provide video results on the \href{http://www.overleaf.com}{project webpage}.

\section{Network and Training}\label{sec:network-and-training}
\subsection{Network Configurations}
\begin{figure}[ht]
  \centering
  \includegraphics[width=1.0\linewidth]{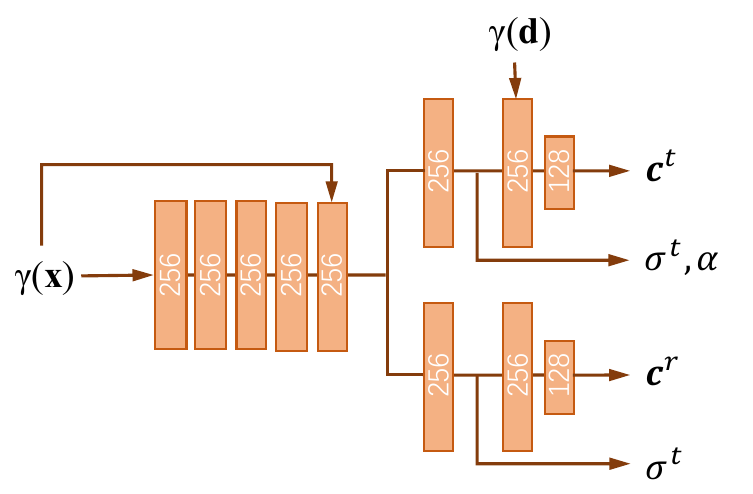}
  \caption{Detailed network architecture of {\sysname}.}
  \label{fig:network-detail}
  \vspace{-0.5cm}
\end{figure}
The detailed network architecture of {\sysname} is shown in \cref{fig:network-detail}. Orange blocks are fully-connected layers. Each layer is followed by ReLU activation except for the output layers. $\gamma(\mathbf{x})$ and $\gamma(\mathbf{d})$ are positional encodings of the input coordinate $\mathbf{x}$ and viewing direction $\mathbf{d}$. The network is designed to have approximately same amount of parameters with the original NeRF~\cite{mildenhall2020nerf} network.

\subsection{Warm-Up Training}
\begin{figure}[ht]
  \centering
  \includegraphics[width=0.9\linewidth]{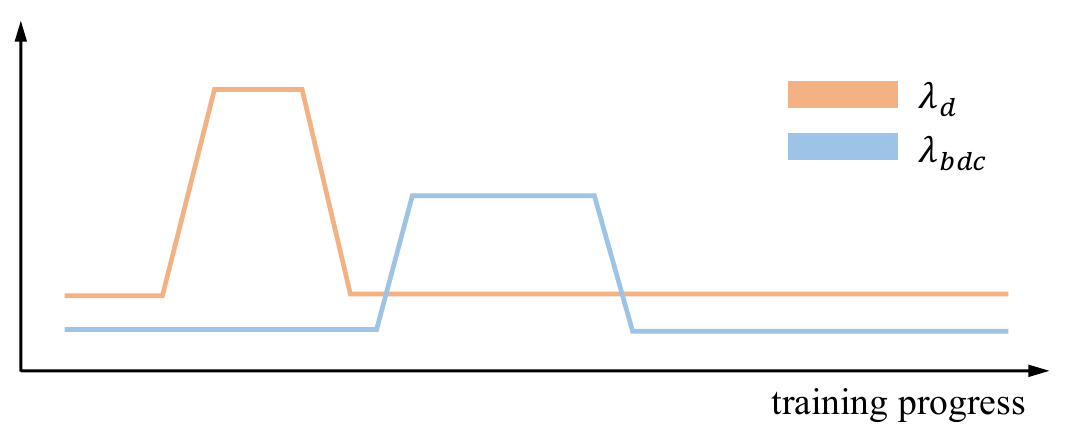}
  \caption{Illustration for the proposed warm-up training. Weightings for geometric priors are first increased then decreased.}
  \label{fig:warm-up}
  \vspace{-0.5cm}
\end{figure}

An illustration of how $\lambda_{d}$ and $\lambda_{bdc}$ change with the training process is shown in \cref{fig:warm-up}.
Note that we first increase $\lambda_{d}$ to ensure correct geometry for the transmitted component and then increase $\lambda_{bdc}$ to remove redundancies in the reflected component. We find this stabilize training compared to optimizing both at the same time. Then we gradually decrease the weights to concern more on photometric loss to get more accurate renderings. Effects of warm-up training are demonstrated in \cref{sec:ablation-warm-up}.

\begin{table*}
  \centering
  \renewcommand\tabcolsep{5pt}
  \begin{tabular}{@{}l|cccccccc|c@{}}
    \toprule
    \multicolumn{1}{c}{} & \multicolumn{9}{c}{PSNR$\uparrow$} \\
    Method & fern & flower & fortress & horns & leaves & orchids & room & trex & average \\
    \midrule
    NeRF & 26.01 & 31.70 & 33.23 & 32.92 & 22.98 & 21.38 & 37.04 & 32.03 & 29.66 \\
    NeRFReN & 25.36 & 29.95 & 33.60 & 31.59 & 22.75 & 21.89 & 37.52 & 30.64 & 29.16 \\
    \bottomrule
  \end{tabular}
  \caption{View synthesis results of NeRF and {\sysname} on LLFF dataset.}
  \label{tab:compare-llff}
\end{table*}

\subsection{Training Details}
We sample $4\times 4$ patches instead of individual pixels to compute the depth smoothness constraint, which requires depth value of pixels in a local area. $\lambda_d$ and $\lambda_{bdc}$ are set to 0.01 and 1e-4 at the beginning of training. $\lambda_d$ is increased to 0.1 at iteration 1k, and decreased to 0.01 at iteration 5k. $\lambda_{bdc}$ is increased to 0.05 at epoch 20, decreased to 1e-4 at epoch 12 and further decreased to 0 at epoch 15. We mask out the input viewing direction before epoch 10. All models are trained for 40 epochs using Adam~\cite{kingma2014adam} optimizer with an initial learning rate of 5e-4. The learning rate is exponentially decayed to 5e-6 in the last 20 epochs.

To further stabilize training, we force the transmitted image to be close to the input image for the first 1k iterations:
\begin{equation}
\mathcal{L}_{init}=||\widehat{\mathbf{C}}(\mathbf{r};\sigma^t,\mathbf{c}^t)-\widehat{C}||_2
\end{equation}
We also apply a smoothness constraint on the reflection fraction map:
\begin{equation}
\mathcal{L}_{\beta}=\sum_{p}\sum_{q\in\mathcal{N}(p)}||\beta(p)-\beta(q)||_1,
\end{equation}
$\lambda_{init}$ and $\lambda_{\beta}$ are set to 0.01 and 1e-4 respectively.

Same as the original NeRF, we simultaneously optimize a coarse and a fine network. 
When training the coarse network, the transmitted and reflected field share the same set of samples. 
For the fine network, we get two different sets of fine samples respectively for the transmitted and reflected field based on the weights of their coarse samples, since the transmitted and reflected component have independent geometries. 
A side-effect would be that we have to evaluate the fine network twice for a query point. 
This computation overhead is carefully taken care of in the comparisons with the original NeRF: 64 fine samples are evaluated for {\sysname} and 128 fine samples for NeRF.

\section{Limitations and Failure Case}\label{sec:failure-case-example}
\begin{figure}[ht]
  \centering
  \includegraphics[width=1.0\linewidth]{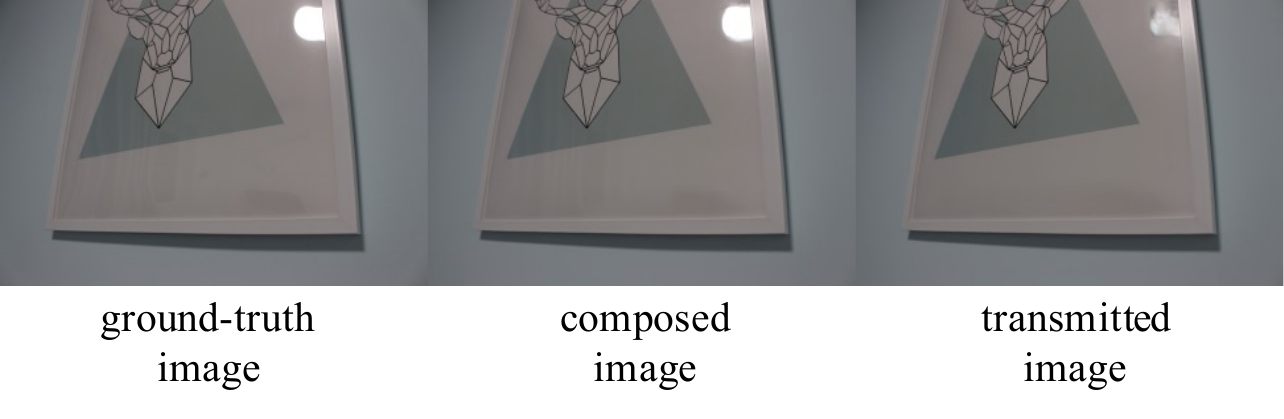}
  \caption{Illustration for a failure case where the virtual image is not stable due to the curved reflective surface. Zoom in for more details.}
  \label{fig:failure}
  \vspace{-0.5cm}
\end{figure}

Curved reflective surfaces that do not produce stable virtual images cannot be modeled by our image-based formulation. An example can be seen in \cref{fig:failure}. Note the distorted light and curtain in the reflections of the ground-truth image. In this case, the reflected component does not have a consistent geometry, leading to inaccurate decomposition and rendering results.

Another limitation lies in the modeling of multiple non-coplanar surfaces. However, we find that in real life, reflection images from different non-coplanar surfaces rarely coincide because they are often far away from each other or only observed from limited angles. This makes it possible to model them by a single reflected field as we do in the paper. Potential failure cases could be alleviated by utilizing multiple reflected MLPs. An interesting direction for addressing the above limitations would be to model reflected rays as in ray tracing, which we regard as a future work.

\section{Results on LLFF Dataset}\label{sec:llff}
\begin{figure}[ht]
  \centering
  \includegraphics[width=1.0\linewidth]{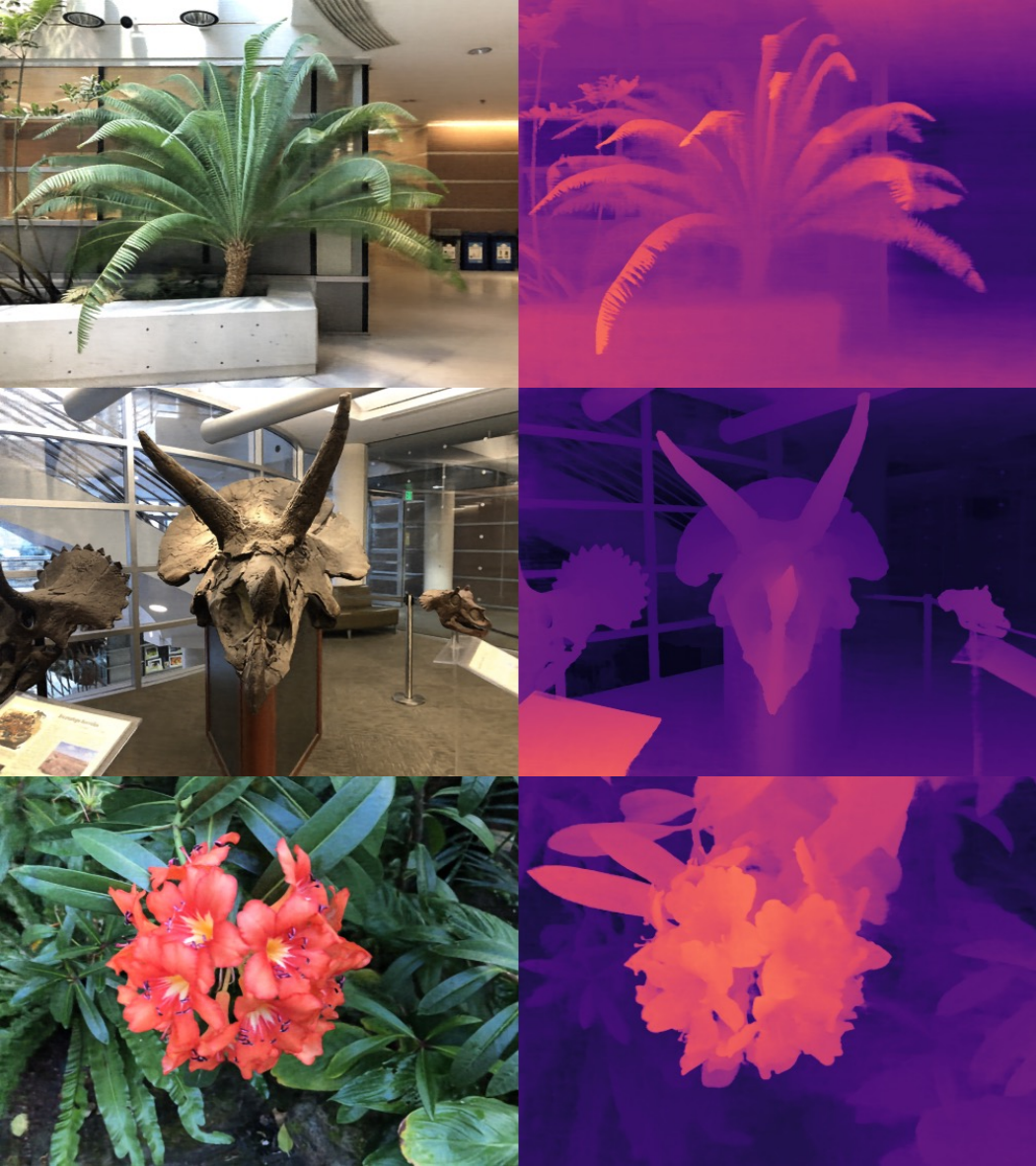}
  \caption{Novel view synthesis and depth estimation results of {\sysname} on some of the LLFF scenes.}
  \label{fig:llff}
  \vspace{-0.5cm}
\end{figure}

\begin{figure*}[t]
  \centering
   \includegraphics[width=1.0\linewidth]{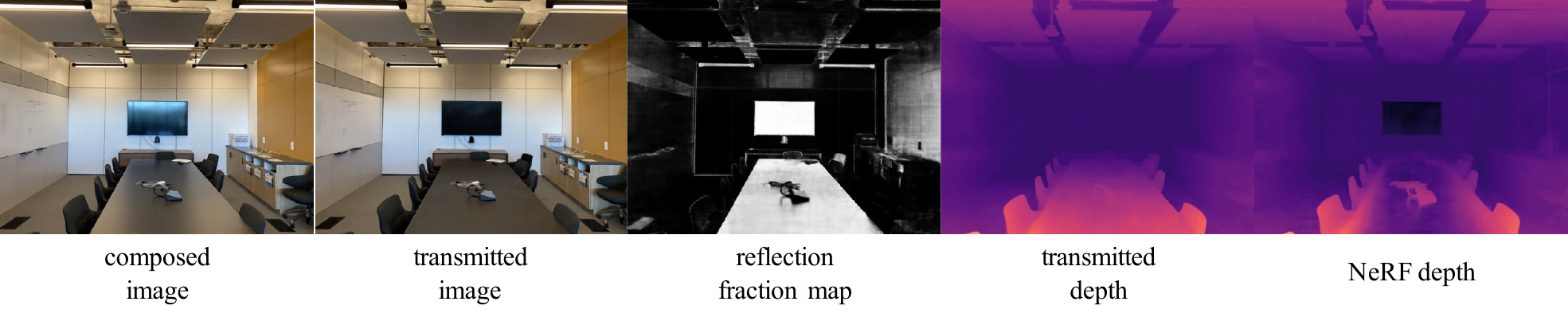}
   \caption{Decomposition results of {\sysname} on the \textit{room} scene with 2 manually annotated reflection masks on the screen. {\sysname} provides reasonable decomposition along with high-quality depth estimation results.}
   \label{fig:llff-room}
   \vspace{-0.5cm}
\end{figure*}

We experiment on LLFF dataset, where most of the scenes do not exhibit strong reflections. As is shown in \cref{tab:compare-llff}, {\sysname} achieves a competitive average PSNR (29.16) compared to NeRF (29.66). This demonstrates that our method maintains high representational ability despite of all the specifically-designed priors in training. Some qualitative results are provided in \cref{fig:llff}. In \cref{fig:llff-room}  we  show  the  decomposition results and improved depth prediction on the \textit{room} scene, where we make use of manually-annotated masks on the TV screen for 2 of the training images. No meaningful decomposition is achieved in other LLFF scenes.

\begin{figure*}[t]
  \centering
  \includegraphics[width=1.0\linewidth]{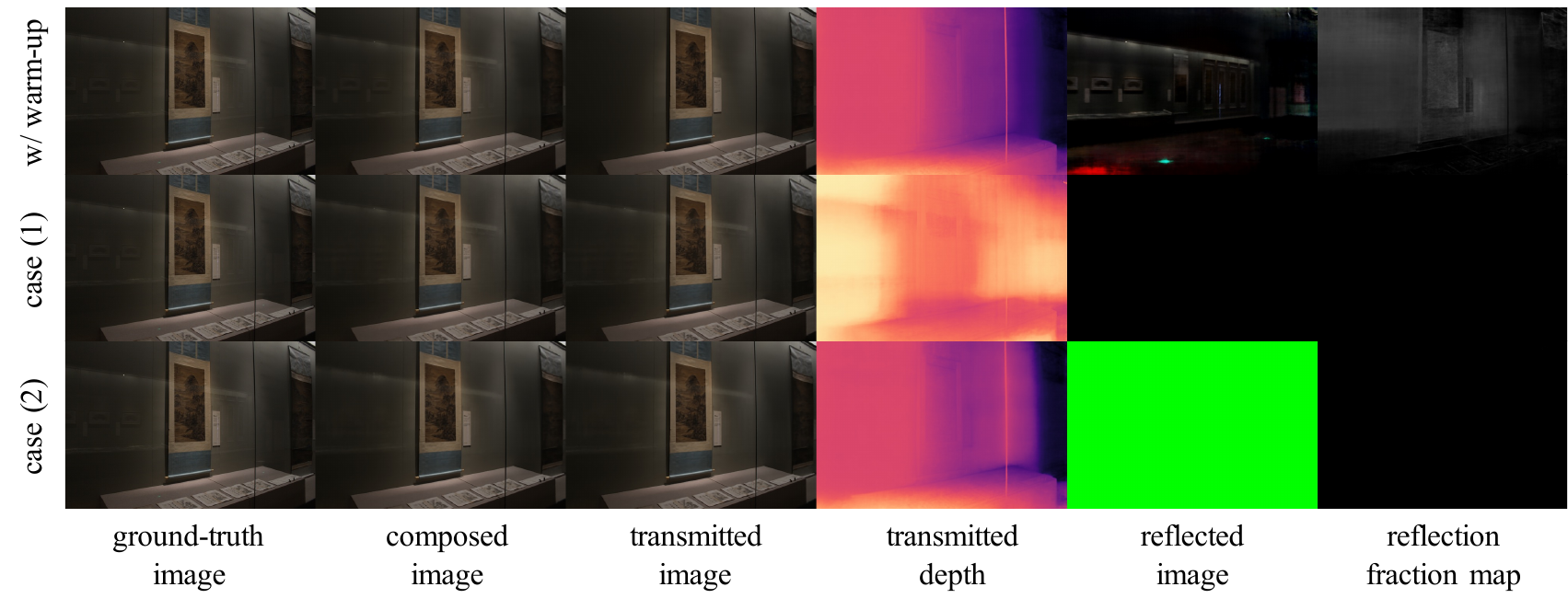}
  \caption{The effects of the proposed warm-up training strategy. The decomposition is likely to fail if strong geometric constraints (row 2) or view-dependency (row 3) are introduced from the beginning of training. We manage to get faithful decomposition by warming-up the weighting factors of the geometric constraints and masking out the viewing direction in early training stage (row 1).}
  \label{fig:ablation-warm-up}
  \vspace{-0.5cm}
\end{figure*}

\section{Ablation Studies}\label{sec:ablation-studies}
\subsection{Warm-up Training}\label{sec:ablation-warm-up}
We exploit two alternative training strategies to demonstrate the effectiveness of warm-up training: 
(1) training with strong geometric constraints ($\mathcal{L}_{d}=0.1, \mathcal{L}_{bdc}=0.05$) directly without warm-up; 
(2) training with viewing directions directly from the very beginning without masking.
The qualitative results are shown in \cref{fig:ablation-warm-up}.
For the first setting, the transmitted depth can be overly smoothed, and the reflected component quickly converges to empty due to the strong constraints without proper initialization. 
For the second setting, the reflected image is explained by view-dependency, leading to blurry reflections lacking fine details. Training with the warm-up strategy does not exhibit such problems, as shown in the first row.

\subsection{Interactive Setting}\label{sec:ablation-interactive}
\begin{figure*}[t]
  \centering
  \includegraphics[width=1.0\linewidth]{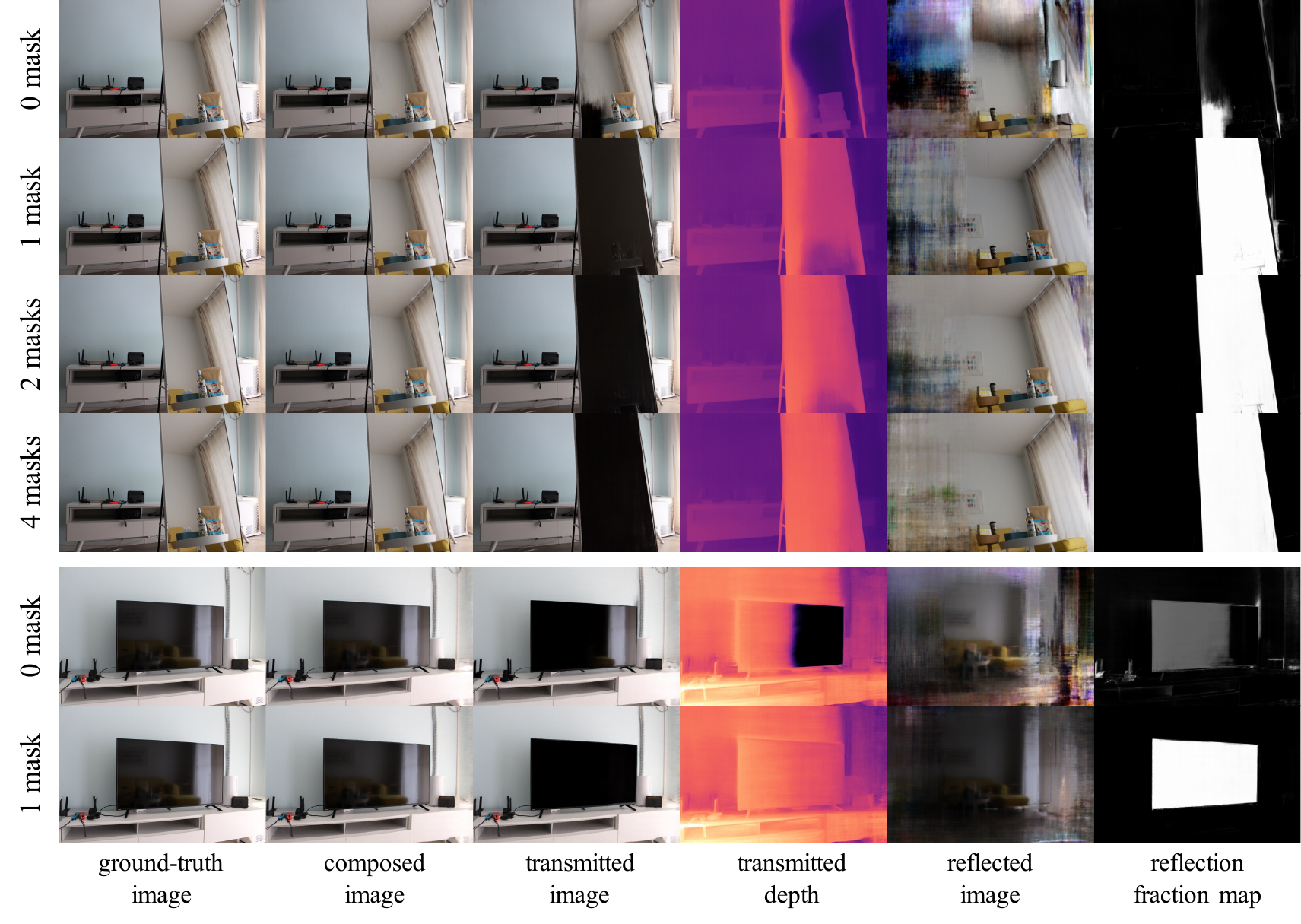}
  \caption{The effects of the user-provided reflection masks on challenging scenes. Without any masks, {\sysname} fails to distinguish between the transmitted and reflected geometry. We utilize 4 masks for the \textit{mirror} scene and only 1 mask for the \textit{tv} scene to get faithful decompositions.}
  \label{fig:ablation-interactive}
  \vspace{-0.5cm}
\end{figure*}

We use user-provided reflection masks to guide the decomposition for the \textit{mirror} and \textit{tv} scene. \cref{fig:ablation-interactive} shows the effects of different numbers of masks to the decomposition results. Without masks, the network finds it hard to distinguish between the transmitted and reflected geometry. For the \textit{mirror} scene, fair decomposition results can be achieved by utilizing 4 masks. And only 1 mask is needed for the \textit{tv} scene. This demonstrates that our method can deal with hard cases with only minimum user inputs.

\end{document}